# Spherical-to-ERP Epipolar Rectification for Single-Axis Disparity in 360° Stereo

[1]Sahereh Obeidavi, [2]Dieter Landes
[1] Faculty of Electrical Engineering and Computer Science, Coburg University of Applied Science, German
[2] Faculty of Electrical Engineering and Computer Science, Coburg University of Applied Science, German
[1]sahereh.obeidavi@hs-coburg.de, [2]dieter.landes@hs-coburg.de

***Abstract—*** ***Omnidirectional stereo images provide full-surround perception but violate the geometric assumptions of classical disparity estimation: in spherical or fisheye views, epipolar correspondences follow curved great-circle paths, producing two-dimensional displacements that cannot be treated as single-axis disparity before geometric rectification. In this work, we adopt a standard spherical-to-equirectangular (ERP) projection as a preprocessing step, which straightens epipolar curves and restores a one-dimensional disparity structure—horizontal for left–right rigs and vertical for top–bottom rigs. Building on our previously introduced RAFT + Epipolar-Aligned Channel Selection (EACS) framework, originally developed for rectilinear and ERP stereo, we examine whether the same modular pipeline remains accurate when the input originates from spherical stereo imagery. After ERP projection, dense optical flow from RAFT is reduced to disparity by retaining only the baseline-aligned flow component. Experiments on synthetic fisheye stereo datasets show that this spherical→ERP→RAFT+EACS pipeline produces accurate, smooth, and structurally consistent disparity maps at real-time speed. These findings confirm that established ERP preprocessing can be effectively combined with our earlier RAFT+EACS method to enable practical, interpretable, and efficient disparity estimation from spherical stereo, providing a straightforward pathway for extending conventional stereo pipelines to 360° imaging.***

## I. INTRODUCTION

Omnidirectional (spherical/fisheye) cameras are increasingly used in robotics, autonomous navigation, and immersive sensing because they provide near-360° coverage with compact hardware [1,2]. This panoramic field of view is particularly valuable for collision avoidance and situational awareness, where blind spots from conventional pinhole cameras can be catastrophic. Yet, the geometric properties of spherical imaging challenge standard stereo pipelines. In rectilinear stereo, disparity is a one-dimensional quantity aligned with the epipolar axis, enabling efficient correspondence search and metrically interpretable depth recovery [3,4]. In contrast, raw spherical pairs exhibit curved epipolar trajectories on the viewing sphere; correspondences in the image plane appear with two non-zero displacement components (horizontal and vertical), even under ideal calibration [5–7]. As a result, disparity is no longer a single-axis shift, and naïvely applying planar disparity or optical-flow magnitude heuristics leads to systematic errors and brittle behavior in low-texture regions.

Recent panoramic depth-estimation studies also emphasize the importance of working in the equirectangular domain. Transformer-based and bi-projection fusion models, such as EGFormer [8] and Elite360D [9], explicitly leverage ERP representations as the primary geometric space for dense prediction. Similarly, dual-projection fusion approaches like GFusion [10] highlight that ERP remains a critical reference projection even when combined with alternative spherical parameterizations.

This effect is illustrated in Figure 1, where spherical stereo introduces two coupled displacement components, but ERP rectification restores the single-axis disparity required for stable correspondence search.

Recent works [9-12], in omnidirectional depth estimation also treats ERP as a stable geometric domain, reinforcing its suitability as a preprocessing step for correspondence-based stereo.

The proposed view is deliberately modular. Rather than learning a monolithic end-to-end network to “internalize” spherical geometry, we externalize the geometry via ERP rectification and then apply standard dense correspondence estimators on the rectified domain. In our reference implementation, dense optical flow (e.g., RAFT) is computed on ERP pairs [13], and disparity is directly read off the epipolar-aligned component of the flow field (horizontal for left–right, vertical for top–bottom). This stands in contrast to using the 2-D flow magnitude or training a dedicated disparity network from scratch [14,15]. The result is a pipeline that is accurate, interpretable, and deployable in real time on commodity GPUs.

This work complements and extends our prior study on rectilinear stereo, which introduced a Epipolar-Aligned Channel Selection (EACS) to project optical flow onto the disparity axis [16]. Here, we tackle the upstream difficulty unique to spherical imaging: before any channel selection can be meaningful, the geometry must be rectified so that epipolar directions are globally consistent. We formalize the fisheye-to-ERP mapping, analyze how epipolar geometry transforms under ERP [11,17], and establish conditions under which the orthogonal flow component ideally vanishes. In doing so, we provide both the theoretical justification and the practical recipe for single-axis disparity extraction from spherical inputs.

The significance of this contribution is twofold. First, it provides a principled bridge between spherical sensing and standard stereo, enabling practitioners to reuse mature correspondence algorithms without bespoke spherical networks [6,14]. Second, by restoring one-dimensional disparity, it improves numerical stability and computational efficiency: correspondence search collapses to a single axis;

memory-intensive 4-D cost volumes become unnecessary [15]; and downstream depth recovery can rely on closed-form spherical triangulation with known intrinsics and baseline [7,12]. Empirically, we show that ERP rectification followed by epipolar-aligned projection yields smooth, structurally consistent disparity maps, preserves thin structures, and remains robust in texture-poor regions—all while sustaining real-time throughput.

Contributions. The paper makes the following contributions:

1) Geometry-aware rectification for spherical stereo. We show that fisheye pairs projected to ERP yield epipolar-aligned scanlines, turning inherently 2-D displacements into single-axis disparities (horizontal or vertical, depending on the rig).
2) Direct disparity extraction in ERP. We demonstrate that, once in ERP, disparity can be obtained by selecting the baseline-aligned flow component and discarding the orthogonal one, enabling lightweight post-processing without retraining.
3) Theory-to-practice pipeline. We detail the mapping, orientation constraints, and disparity estimation for omnidirectional rigs, and we report quantitative/qualitative results on synthetic spherical data that validate accuracy and real-time feasibility.
4) Modularity and deployability. The approach functions as a drop-in preprocessing stage for existing correspondence estimators (including but not limited to RAFT), facilitating integration in robotics stacks where latency, interpretability, and resource budgets are critical.

Our goal is to determine whether explicit geometric rectification enables standard correspondence methods to operate correctly on spherical stereo inputs.

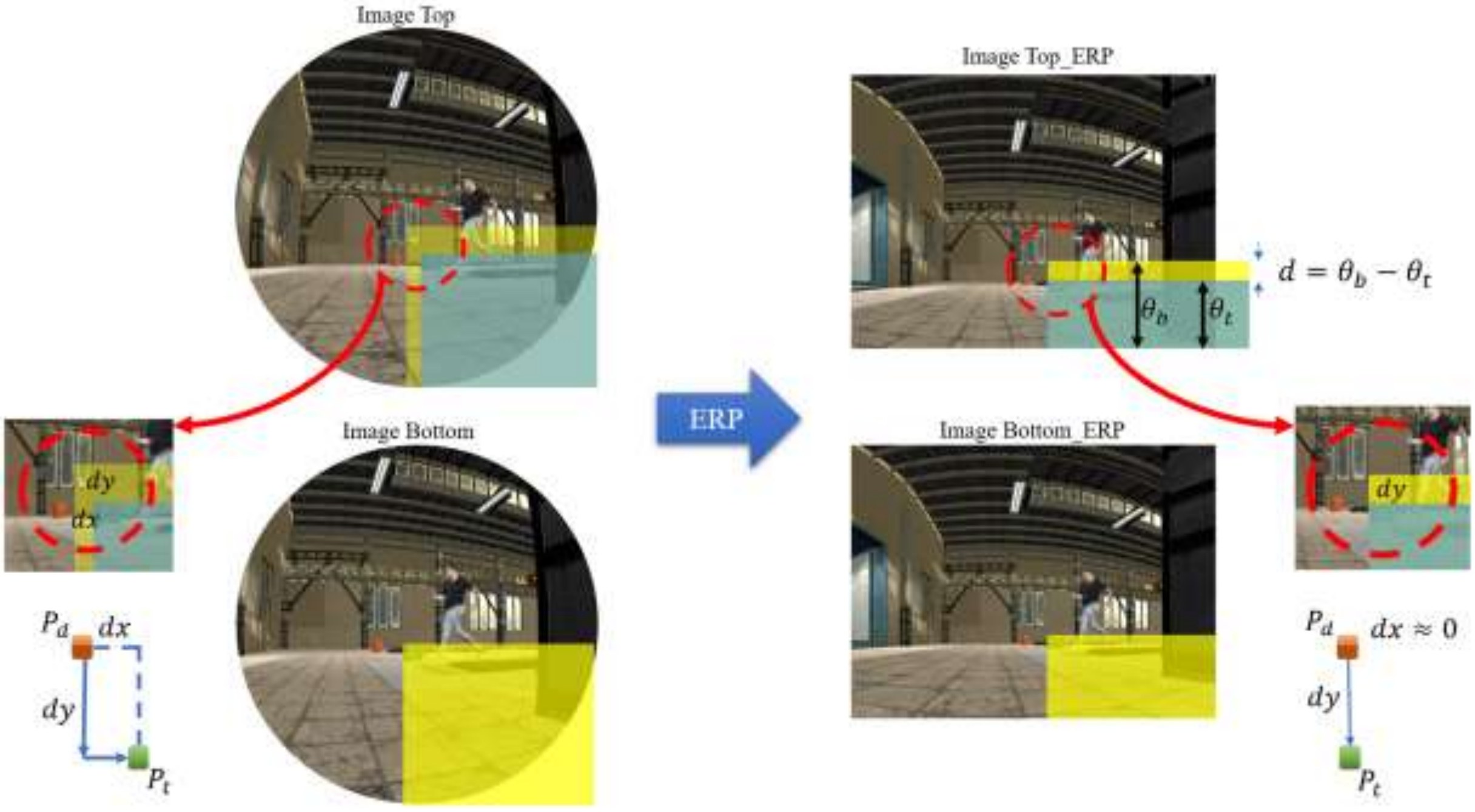

Figure 1. Spherical→ERP rectification restores single-axis disparity for 360° stereo. Left: In the native spherical domain (top/bottom cameras), a patch exhibits coupled horizontal and vertical displacements (dx,dy) along a great-circle epipolar curve. Right: After spherical epipolar rectification and equirectangular projection (ERP), the same patch aligns to the image axis. For a top–bottom rig, the correspondence is vertical only, with pixel disparity dy proportional to the angular offset $\theta_b - \theta_t$, while $dx \approx 0$. For a left–right rig, the situation is analogous with horizontal disparity. This reduction to a one-dimensional displacement enables direct disparity extraction from optical flow via EACS (Paper 1), retaining $f_y$ for top–bottom or $f_x$ for left–right.

## II. Related Work

### A. Stereo Disparity Estimation in Rectilinear Images

Stereo disparity estimation has been extensively studied in computer vision. Early classical methods relied on local correlation windows and global optimization frameworks [4,18]. While conceptually simple, these approaches often failed in regions of low texture, occlusions, or radiometric variation. With the rise of deep learning, convolutional neural networks (CNNs) revolutionized stereo by introducing end-to-end disparity regression. GC-Net [19] and PSMNet [15] pioneered 3D cost volume aggregation, while subsequent refinements such as GA-Net [20] and CREStereo [21] improved both accuracy and efficiency. Benchmarks such as KITTI [22] established rectilinear stereo as a mature research area.

Despite these advances, most networks assume rectified pinhole geometry, where disparity is strictly aligned with a single image axis. This assumption does not hold in spherical imaging, motivating geometric preprocessing before disparity estimation can be meaningfully applied.

### B. Optical Flow and Disparity Connections

Optical flow and stereo disparity are inherently related: both describe dense correspondences between images, but with different geometric constraints. Flow represents general 2D displacements between temporally adjacent frames [23, 24], while disparity in rectified stereo corresponds to a 1D shift along the epipolar axis [3]. Recent methods such as RAFT [13] have set new standards for optical flow accuracy, while RAFT-Stereo [14] adapts the same architecture for disparity. Our prior work, Epipolar-Aligned Channel Selection (EACS) [16], extracts disparity directly from the axis-aligned component of RAFT-generated optical flow in rectilinear stereo. This approach eliminates redundant flow channels and shows that disparity estimation does not require retraining a dedicated stereo network. However, EACS assumes that disparity is already axis-aligned—a condition violated in spherical inputs.

### C. Epipolar Geometry in Spherical Stereo

Fisheye and omnidirectional cameras are increasingly used in robotics and VR/AR applications for their wide fields of view [25–27]. However, in spherical imaging, epipolar curves are no longer straight lines but great-circle arcs, causing correspondences to exhibit displacements in both the horizontal and vertical directions [26,27]. This destroys the 1D disparity assumption and complicates correspondence search.

To mitigate this, projection schemes such as cube maps, gnomonic projection, and Equirectangular Projection (ERP) have been proposed [28, 29]. ERP is particularly appealing because it maps great-circle epipolar curves to straight scanlines, restoring the 1D alignment of disparity. Elhashash and Casper [30] and Müller et al. [31] showed that ERP rectification enables the reuse of standard stereo pipelines, while Pathak and Nishino [32] highlighted the importance of aligning epipolar geometry before applying correspondence algorithms. Beyond stereo, several recent omnidirectional depth-estimation frameworks reaffirm the relevance of ERP as a core projection. EGFormer [8] introduces an ERP-geometry-biased transformer that embeds spherical distortion directly into local attention. Elite360D [9] combines ERP features with icosahedral point encodings for global-local bi-projection fusion. GFusion [10] similarly performs cross-domain fusion between ERP and icosahedral representations. Although these methods focus on monocular or fused depth prediction rather than stereo geometry, they consistently treat ERP as a stable reference domain for 360° scene understanding, further supporting the motivation for ERP-based rectification in stereo pipelines.

### D. Omnidirectional Projections and ERP Alternatives

Recent research also explores alternative spherical projections for tasks such as depth estimation and omnidirectional video processing. GFusion [10] proposes an ERP–ICOSAP fusion strategy to mitigate projection discontinuities in 360° depth estimation. Elite360D [9] likewise adopts a bi-projection architecture combining ERP with an icosahedral point-set representation. For efficient omnidirectional video compression, Filipe et al. [35] introduce a packed Craster–Parabolic projection derived from Collignon geometry. While these approaches address challenges in panoramic representation, they operate primarily at the feature-learning or compression level and do not enforce epipolar alignment for stereo correspondence. In contrast, our work focuses on geometric rectification that restores the one-dimensional nature of disparity.

### E. Summary and Gap

Stereo disparity estimation is well established for rectilinear images [4, 18–22], and prior work has shown that optical flow contains disparity information when epipolar alignment is enforced [13,16]. Spherical imaging, however, violates this assumption: disparities appear along both image axes simultaneously, breaking the one-dimensional structure required by classical stereo pipelines. Existing studies confirm that equirectangular projection can restore epipolar alignment for spherical cameras [28–32], but they do not investigate whether such rectified representations are compatible with optical-flow–based disparity extraction.

This gap motivates the focus of the present work: to determine whether a standard spherical-to-ERP rectification step provides a representation in which the RAFT + Epipolar-Aligned Channel Selection (EACS) pipeline—originally developed for rectilinear and ERP stereo—can accurately recover single-axis disparity without retraining or designing specialized spherical stereo networks. By examining this compatibility, we evaluate whether dense optical flow combined with EACS remains a viable and efficient approach for disparity estimation when the input originates from spherical stereo imagery. To our knowledge, no prior work combines spherical-to-ERP rectification with dense optical flow and channel-aligned disparity extraction.

## III. Methodology

This work investigates whether disparity can be reliably extracted from spherical stereo imagery by combining a standard spherical-to-equirectangular projection with the previously introduced RAFT + Epipolar-Aligned Channel Selection (EACS) framework. Spherical cameras violate the one-dimensional disparity assumption required by classical stereo, as correspondences exhibit coupled horizontal and vertical displacements. To obtain a domain in which single-axis disparity becomes meaningful, we first project the fisheye inputs into the Equirectangular Projection (ERP) domain. Dense optical flow is then estimated on the rectified ERP pair, and disparity is obtained by retaining only the epipolar-aligned flow component.

### A. Disparity in Spherical Stereo

Consider a pair of fisheye cameras capturing images on the unit sphere. A 3D point $P$ is mapped to angular coordinates $(\theta_l, \theta_r)$ in the left and right spherical images, respectively. Following a standard spherical camera model [34], the disparity between these two projections may be expressed as an angular offset:

$$d_s = f_s(\theta_l - \theta_r) \tag{1}$$

Where $f_s$ is the radius of the great circle (equivalently, the focal length under a spherical camera model). For normalized spherical images ($f_s$ =1), the disparity simplifies to the angular difference:

$$d_n = \theta_l - \theta_r \tag{2}$$

Unlike rectilinear disparity, which is strictly horizontal in a rectified pinhole stereo pair, spherical disparity generally has two components (longitude and latitude), yielding 2D displacements in the image plane. This dual displacement makes disparity extraction unstable if performed directly in the spherical domain.

### B. Rectification via Equirectangular Projection (ERP)

To recover a domain in which disparity aligns with a single image axis, we employ the Equirectangular Projection (ERP) as a geometric preprocessing step. ERP maps spherical coordinates $(\varphi, \theta)$, where $\varphi$ is longitude and $\theta$ is latitude, to a 2D grid:

$$u = f_\varphi \varphi, \qquad v = f_\theta \theta \tag{3}$$

with $f_\varphi$ and $f_\theta$ representing the horizontal and vertical scaling factors (in pixels per radian). Under ERP, great circles that define spherical epipolar curves are transformed into straight lines aligned with the grid axes [29, 31, 34]. By orienting the baseline appropriately:

- Left–right fisheye pairs yield purely horizontal disparities (Δu),
- Top–bottom fisheye pairs yield purely vertical disparities (Δv).
- This rectification removes the mixed displacement components that characterize the spherical image plane and restores the one-dimensional disparity structure required by correspondence-based stereo algorithms. Figure 2 illustrates the transformation from fisheye images to ERP and the resulting alignment of epipolar geometry.

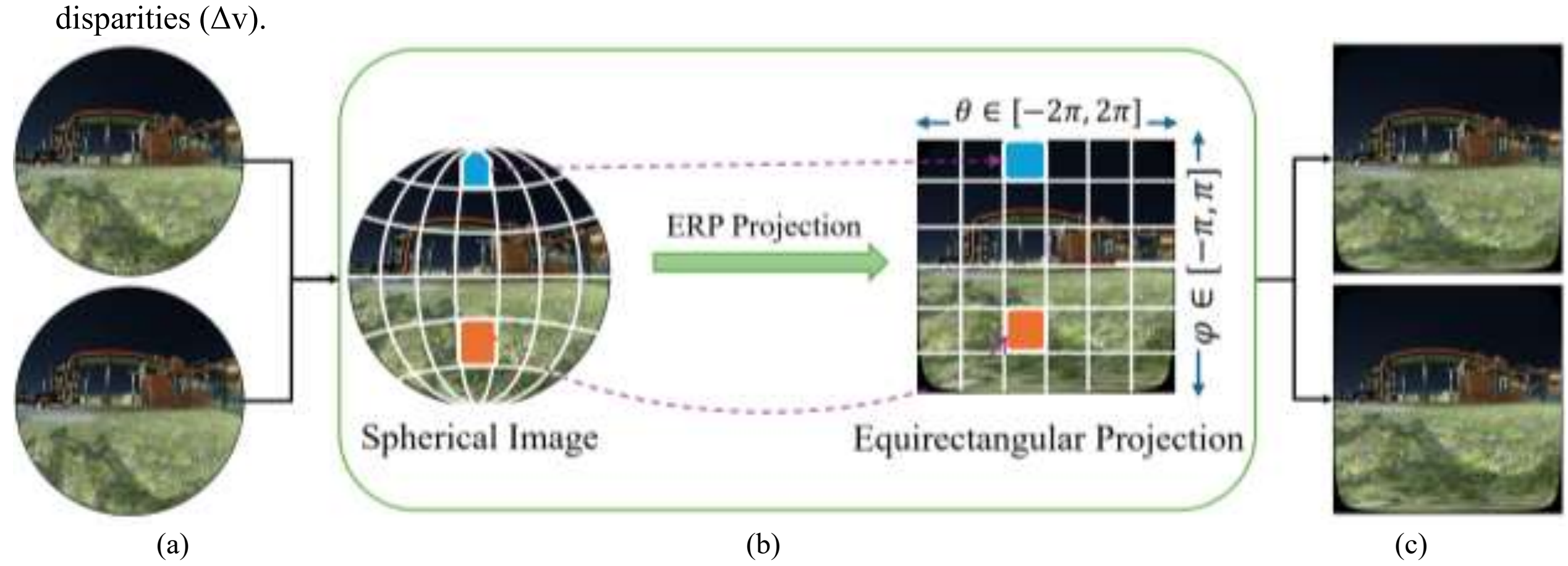

(a) (b) (c)

Figure 2 — Transformation of fisheye images into the equirectangular projection (ERP) domain. (a) Original fisheye images captured by the omnidirectional stereo rig. (b) Ray-based reprojection onto the unit sphere, producing the spherical image parameterized by azimuth $\theta \in [-2\pi, 2\pi]$ and elevation $\varphi \in [-\pi, \pi]$. (c) Equirectangular projection obtained through latitude–longitude sampling, where spherical coordinates are mapped to a 2D grid. Highlighted colored patches demonstrate the correspondence between regions on the sphere and their locations in the ERP image. In this domain, corresponding projections of the same 3D point appear as two points aligned on the same horizontal line (scanline), with their horizontal separation representing the azimuthal disparity $d_\theta$. This property enables efficient stereo correspondence search along horizontal scanlines in ERP space.

*C. Flow-Based Disparity Extraction Using RAFT + EACS*

Once ERP rectification is applied, dense correspondence can be estimated using a general optical flow network. In this work, we em After ERP rectification, dense correspondences are computed usingRAFT [13],
which estimates a two-channel flow field:

$$F(u,v) = (f_x(u,v), f_y(u,v)). \quad (4)$$

Because ERP enforces a single epipolar direction, only one of these channels corresponds to true disparity. Therefore, we apply Epipolar-Aligned Channel Selection (EACS) [16]:

- For left–right ERP pairs: $d(u,v) = f_x(u,v)$.
- For top–bottom ERP pairs: $d(u,v) = f_y(u,v)$.

This projection discards the irrelevant
channel, retaining only the axis consistent with the stereo baseline. This operation yields a single disparity map directly from the optical flow.

*D. Pipeline Overview*

Figure 3 summarizes the full process, from spherical imagecapture to ERP rectification, optical-flow estimation, and disparity extraction.

The proposed disparity estimation pipeline consists of:

1) Capture stereo fisheye/spherical image pairs.
2) Project both images into the ERP domain.
3) Compute dense optical flow between the ERP pair using RAFT.
4) Apply EACS to select the epipolar-aligned component of flow as disparity.
5) Output disparity maps suitable for downstream processing.

By combining ERP rectification with flow-based EACS, we achieve a modular, geometry-aware approach that restores the definition of disparity for spherical stereo. Unlike prior methods that train specialized spherical disparity networks, our pipeline leverages existing correspondence estimators and lightweight post-processing, making it both interpretable and computationally efficient. The proposed pipeline is summarized in Figure 3, which shows how spherical stereo inputs are rectified into ERP, processed by RAFT to estimate dense correspondences, and then converted to disparity maps via EACS.

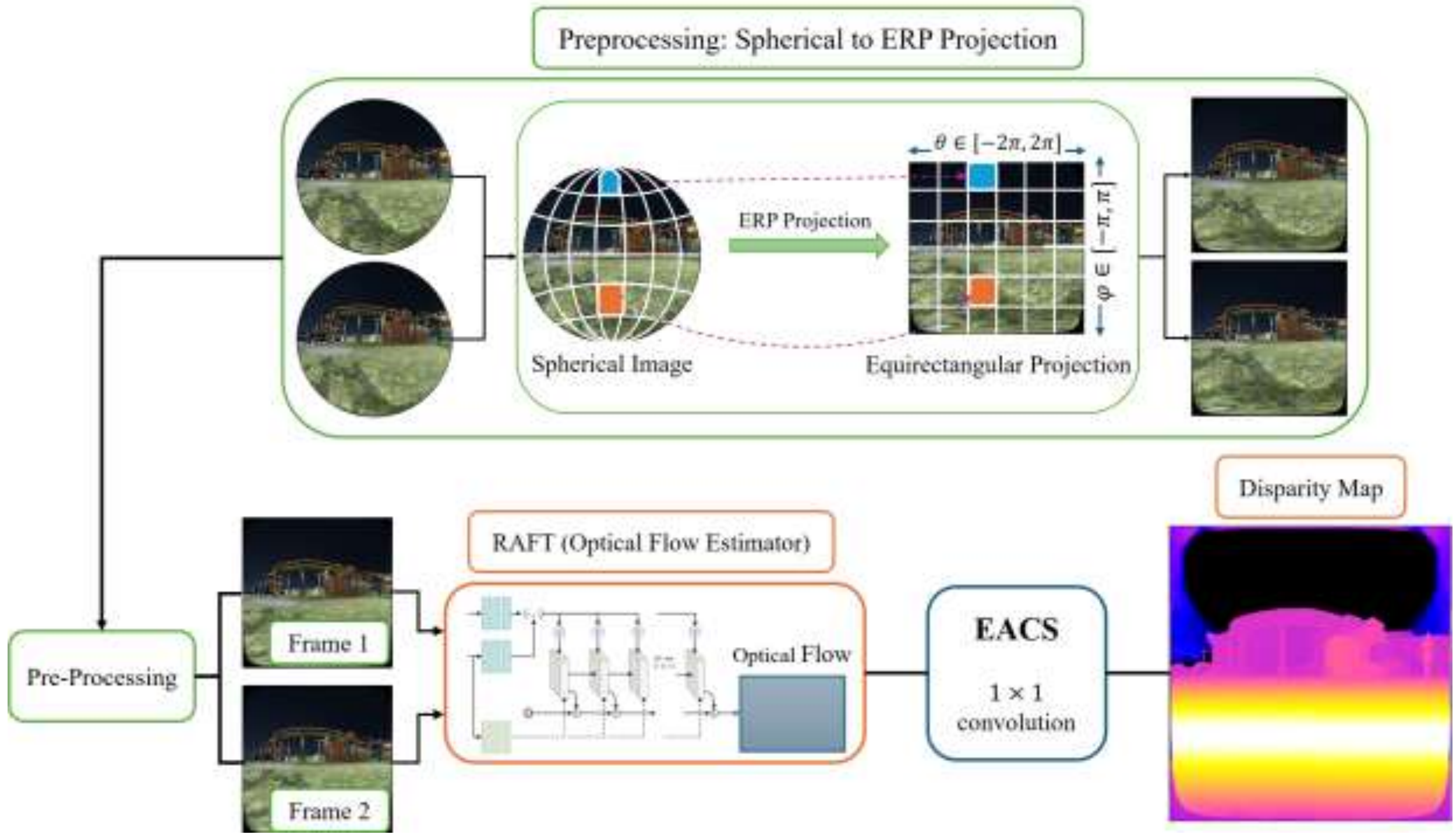

Figure 3. Overview of the proposed spherical stereo disparity estimation pipeline. In the preprocessing stage, fisheye stereo images are reprojected into the equirectangular projection (ERP) domain, where epipolar curves align with image scanlines. The rectified ERP image pairs are then passed to a RAFT-based optical flow estimator [13], which produces dense 2D correspondences. Epipolar-Aligned Channel Selection (EACS) [16] projects the flow onto the baseline-aligned axis (horizontal for left–right rigs, vertical for top–bottom rigs), yielding the final single-axis disparity map.

## IV. Experiments

The experimental study evaluates whether ERP rectification enables the RAFT + Epipolar-Aligned Channel Selection (EACS) configuration to extract disparity from spherical stereo inputs without architectural modification, and how its performance compares to established RAFT-based stereo approaches when operating on the same ERP-rectified data. For this purpose, a synthetic omnidirectional stereo dataset generated at TU Chemnitz was employed. Each sample comprises fisheye stereo RGB views together with dense ground-truth disparity and depth annotations. All fisheye images were reprojected to the equirectangular domain at 1600×1600 resolution using longitude–latitude sampling, thereby providing identical input formatting for all compared methods [34].

The evaluation addresses two objectives. The first is to determine whether ERP rectification provides a representation in which disparity can be extracted as a single-axis displacement using RAFT [13] combined with EACS [16] without retraining or architectural modification. In order to position to contextualize the resulting performance against established RAFT-derived stereo pipelines—CREStereo [21], RAFT-Stereo [14], and LEAStereo [35]—which extend RAFT flow reasoning toward dense disparity computation. These methods were therefore selected for comparison because their objective (dense correspondence and disparity estimation) and foundational architecture (RAFT-based refinement) render their accuracy and runtime metrics appropriate comparative dimensions under identical ERP-rectified inputs.

All methods were evaluated on the same dataset and the same ERP-rectified stereo inputs. RAFT [13] was applied to compute dense correspondence fields, after which EACS [16] was used to retain only the flow channel aligned with the stereo axis while discarding the orthogonal component. No fine-tuning or domain adaptation was performed for any method. Runtime measurements were carried out on an NVIDIA TU102 GPU at full ERP resolution and include projection overhead. Disparity accuracy was quantified using mean absolute error (MAE) and root mean squared error (RMSE), computed against the provided synthetic ground truth. These metrics were selected because they directly characterize the deviation between predicted and true disparity magnitudes across the depth range present in the omnidirectional scenes.

Disparity accuracy is quantified using mean absolute error (MAE) and mean squared error (MSE) between predicted and ground-truth disparity. Furthermore, the D1-all outlier rate and the proportions of pixels with errors greater than 3 and 5 pixels (>3 px, >5 px) are reported, following common stereo evaluation practice [4, 21, 22]. These complementary measures characterise both average error magnitude and robustness to larger deviations. Since the intended application domain includes real-time robotic and omnidirectional perception, runtime in frames per second (FPS) is treated as a primary criterion alongside accuracy.

Table 1 summarises the quantitative comparison. On the ERP-rectified spherical dataset, the ERP + RAFT + EACS pipeline attains a MAE of 0.3007 and an MSE of 0.9470, with D1-all, >3 px, and >5 px outlier rates of 0.40%, 0.30%, and 0.05%, respectively, at a throughput of 10 fps. CREStereo [21] exhibits a higher MAE (0.3581) and D1-all outlier rate (0.98%), together with a lower runtime of 3.5 fps. RAFT-Stereo [14] and LEAStereo [35] report D1-all outlier rates of 2.45% and 2.79%, respectively, with runtimes of 5.0 fps and 2.0 FPS on their original rectilinear benchmarks. These figures, collectively illustrate typical accuracy–runtime trade-offs among contemporary stereo networks and show that the proposed ERP + RAFT + EACS configuration occupies a favourable region of this spectrum for ERP-rectified spherical data.

Table 1. Quantitative comparison of our ERP + RAFT + EACS pipeline with representative stereo methods. Accuracy metrics are reported as mean absolute error (MAE), mean squared error (MSE), and outlier rates (D1-all, >3 px, >5 px), where lower is better. Runtime is measured in frames per second (FPS), where higher is better.

| Method | MAE ↓ | MSE ↓ | D1-all (%) ↓ | >3 px (%) ↓ | >5 px (%) ↓ | FPS ↑ |
|---|---|---|---|---|---|---|
| **Ours (ERP + RAFT + EACS)** | **0.3007** | **0.9470** | **0.40** | **0.30** | **0.05** | **10.0** |
| CREStereo [21] | 0.3581 | – | 0.98 | – | 0.21 | 3.5 |
| RAFT-Stereo [14] | – | – | 2.45 | – | 0.36 | 5.0 |
| LEAStereo [35] | – | – | 2.79 | – | 0.46 | 2.0 |

In addition to the quantitative evaluation, qualitative assessment was carried out on the ERP-rectified dataset by comparing predicted disparity maps with the corresponding ground truth. Figure 4 shows a representative example. The ground-truth disparity is depicted in panel (a), the disparity estimated by the ERP + RAFT + EACS pipeline in panel (b), and the pixelwise RMSE error map in panel (c). The error distribution indicates that most regions exhibit small deviations, with noticeable errors concentrated at depth discontinuities and in regions with undefined disparity, such as the sky. Structural contours and thin objects remain well preserved, which is consistent with the low outlier rates reported in Table 1.

Taken together, the results demonstrate that ERP rectification provides a suitable representation for applying RAFT-based dense correspondence to spherical stereo and that EACS enables direct single-axis disparity extraction without additional training. The combination yields competitive accuracy, low outlier rates, and a runtime of 10 fps at full ERP resolution, making the approach attractive for real-time omnidirectional perception scenarios. The comparison with CREStereo [21], RAFT-Stereo [14], and LEAStereo [35] situates the proposed method within the broader class of stereo networks and clarifies its position in terms of the trade-off between disparity accuracy and computational efficiency.

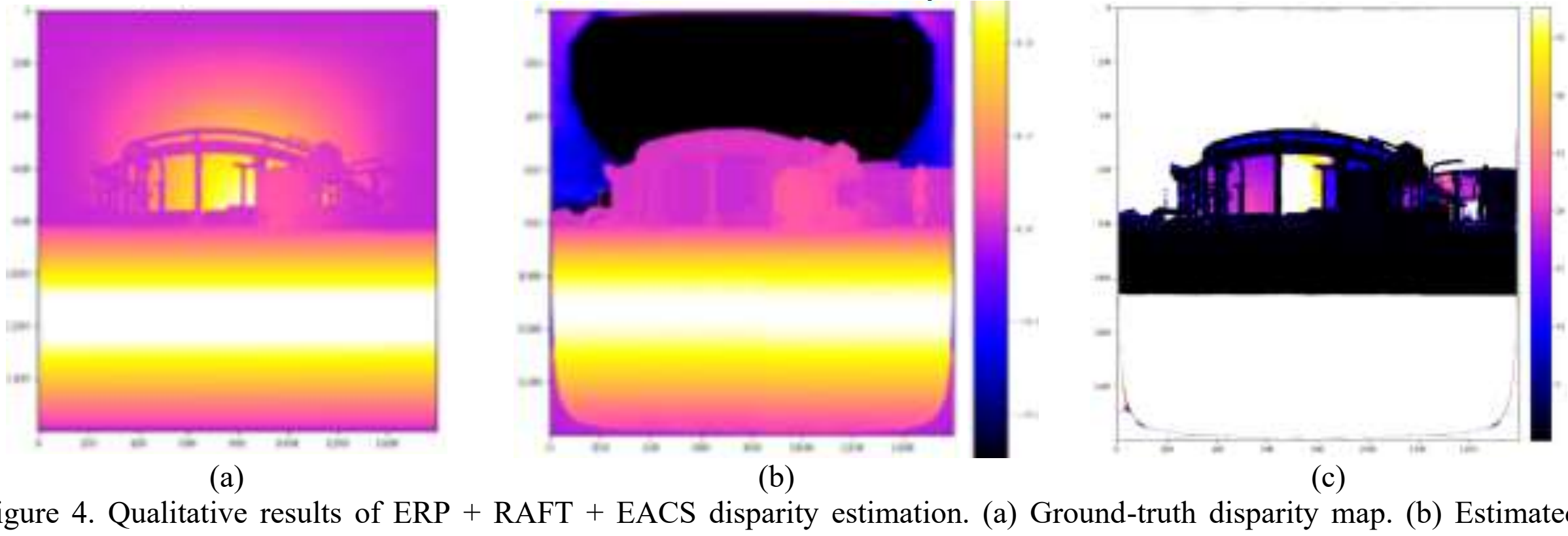

Figure 4. Qualitative results of ERP + RAFT + EACS disparity estimation. (a) Ground-truth disparity map. (b) Estimated disparity map, with undefined sky regions shown in black. (c) Pixelwise RMSE error map, where darker colors indicate smaller errors. The results demonstrate that the proposed method produces disparity maps highly consistent with ground truth, with errors primarily confined to object boundaries.

## CONCLUSION

This work examined the practical feasibility of extracting disparity from spherical stereo imagery by first mapping fisheye inputs into the equirectangular domain and subsequently applying an unmodified RAFT + EACS [16] pipeline. The geometric role of ERP rectification is well documented in prior literature; therefore, the central contribution here was not to reformulate the projection model, but to assess whether the rectified representation is sufficient for single-axis disparity extraction.

Evaluation on synthetic omnidirectional stereo data demonstrated that ERP-rectified inputs allow RAFT + EACS [16] to produce disparity estimates consistent with ground truth while avoiding cross-axis displacement effects inherent to raw spherical correspondences. Furthermore, error metrics and runtime comparisons against RAFT-derived stereo networks indicate that the proposed configuration attains competitive disparity quality while maintaining real-time throughput.

Future extensions include integrating depth reconstruction via spherical triangulation, assessing performance under wider baselines and real-world omnidirectional capture conditions, and investigating alternative projection schemes in cases where ERP distortion characteristics may be suboptimal.

## REFERENCES

[1] Scaramuzza, D., Martinelli, A., & Siegwart, R. (2006). A flexible technique for accurate omnidirectional camera calibration and structure from motion. Proceedings of the IEEE International Conference on Computer Vision (ICCV).

[2] Baker, S., & Nayar, S. (1999). A theory of single-viewpoint catadioptric image formation. International Journal of Computer Vision (IJCV), 35(2), 175–196.

[3] Hartley, R., & Zisserman, A. (2004). Multiple View Geometry in Computer Vision. Cambridge University Press.

[4] Scharstein, D., & Szeliski, R. (2002). A taxonomy and evaluation of dense two-frame stereo correspondence algorithms. International Journal of Computer Vision (IJCV), 47(1–3), 7–42.

[5] Sturm, P., Ramalingam, S., Tardif, J. P., Gasparini, S., & Barreto, J. (2011). Camera models and fundamental concepts used in geometric computer vision. Foundations and Trends in Computer Graphics and Vision, 6(1–2), 1–183.

[6] Poggi, M., Tosi, F., Batsos, K., Mordohai, P., & Mattoccia, S. (2021). On the synergies between machine learning and binocular stereo for depth estimation: A survey. IEEE Transactions on Pattern Analysis and Machine Intelligence (TPAMI).

[7] Elhashash, M., & Casper, R. (2022). Investigating spherical epipolar rectification for multi-view stereo. arXiv preprint, arXiv:2204.04141.

[8] Yun, I., Shin, C., Lee, H., Lee, H.-J., & Rhee, C. E. (2023). Egformer: Equirectangular geometry-biased transformer for 360 depth estimation. Proceedings of the IEEE/CVF International Conference on Computer Vision (ICCV), 6101–6112.

[9] Ai, H., & Wang, L. (2024). Elite360D: Toward efficient 360 depth estimation via semantic- and distance-aware bi-projection fusion. Proceedings of the IEEE/CVF Conference on Computer Vision and Pattern Recognition (CVPR), 9926–9935.

[10] Ma, C., Shen, J., Tie, Y., Zhao, Z., Zhao, J., & Zhang, D. (2025). GFusion: Global Geometry Aware Dual projection fusion for 360 depth estimation. Proceedings of the IEEE International Conference on Electronic Information Technology (EIT), 841–847.

[11] Jiang, S., Xie, Y., Lei, Z., & Wang, Y. (2023). 3D reconstruction of spherical images based on ERP. arXiv preprint, arXiv:2306.12770.

[12] Pathak, S., Moro, A., Yamashita, A., & Asama, H. (2016). Dense 3D reconstruction from two spherical images via optical flow-based ERP epipolar rectification. Proceedings of the IEEE International Conference on Imaging Systems and Techniques (IST), 140–145.

[13] Teed, Z., & Deng, J. (2020). RAFT: Recurrent All-Pairs Field Transforms for Optical Flow. Proceedings of the European Conference on Computer Vision (ECCV).

[14] Lipson, L., Teed, Z., & Deng, J. (2021). RAFT-Stereo: Multilevel Recurrent Field Transforms for Stereo Matching. Proceedings of the International Conference on 3D Vision (3DV).

[15] Chang, J. R., & Chen, Y. S. (2018). Pyramid stereo matching network. Proceedings of the IEEE/CVF Conference on Computer Vision and Pattern Recognition (CVPR).

[16] Obeidavi, S., Landes, D., Moosavipoor M. Epipolar-Aligned Channel Selection: A Projection from Optical Flow to Disparity. International Journal of Computer Applications. 187, 69 ( Dec 2025), 7-16. DOI=10.5120/ijca2025926148.

[17] Kang, D., Jang, H., Lee, J., Kyung, C.-M., & Kim, M. H. (2022). Uniform subdivision of omnidirectional camera space for efficient spherical stereo matching. Proceedings of the IEEE/CVF Conference on Computer Vision and Pattern Recognition (CVPR), 12962–12970.

[18] Hirschmüller, H. (2008). Stereo processing by semiglobal matching and mutual information. IEEE Transactions on Pattern Analysis and Machine Intelligence (TPAMI), 30(2), 328–341.

[19] Kendall, A., et al. (2017). End-to-end learning of geometry and context for deep stereo regression. Proceedings of the IEEE International Conference on Computer Vision (ICCV).

[20] Zhang, F., et al. (2019). GA-Net: Guided aggregation net for end-to-end stereo matching. Proceedings of the IEEE/CVF Conference on Computer Vision and Pattern Recognition (CVPR).

[21] Li, J., et al. (2022). Practical stereo matching via cascaded recurrent network with adaptive correlation. Proceedings of the IEEE/CVF Conference on Computer Vision and Pattern Recognition (CVPR).

[22] Geiger, A., et al. (2013). Vision meets robotics: The KITTI dataset. International Journal of Robotics Research (IJRR), 32(11), 1231–1237.

[23] Horn, B., & Schunck, B. (1981). Determining optical flow. Artificial Intelligence, 17(1–3), 185–203.

[24] Sun, D., Yang, X., Liu, M.-Y., & Kautz, J. (2018). PWC-Net: CNNs for optical flow using pyramid, warping, and cost volume. Proceedings of the IEEE/CVF Conference on Computer Vision and Pattern Recognition (CVPR).

[25] Baker, S., & Nayar, S. (1998). A theory of catadioptric image formation. Proceedings of the IEEE International Conference on Computer Vision (ICCV), 35–42.

[26] Barreto, J., & Araújo, H. (2001). Issues on the geometry of central catadioptric image formation. Proceedings of the IEEE/CVF Conference on Computer Vision and Pattern Recognition (CVPR).

[27] Geyer, C., & Daniilidis, K. (2000). A unifying theory for central panoramic systems. Proceedings of the European Conference on Computer Vision (ECCV).

[28] Zioulis, N., Karakottas, A., Zarpalas, D., & Daras, P. (2018). OmniDepth: Dense depth estimation for indoor spherical panoramas. Proceedings of the European Conference on Computer Vision (ECCV).

[29] Coors, R., Condurache, A. P., & Geiger, A. (2018). SphereNet: Learning spherical representations for detection and classification in omnidirectional images. Proceedings of the European Conference on Computer Vision (ECCV).

[30] Elhashash, Y., & Casper, J. (2022). Stereo matching rectification for spherical images using ERP. Proceedings of the IEEE International Conference on Image Processing (ICIP).

[31] Müller, K., Kuhn, F., & Koch, R. (2019). Robust stereo estimation for spherical images. Computer Graphics Forum, 38(2), 59–71.

[32] Pathak, D., & Nishino, K. (2017). Learning socially aware person representation for 360° videos. Proceedings of the IEEE/CVF Conference on Computer Vision and Pattern Recognition (CVPR).

[33] Filipe, J. N., Tavora, L. M. N., de Faria, S. M. M., Navarro, A., & Assuncao, P. A. A. (2025). An efficient projection for fast low bitrate omnidirectional video compression. Proceedings of the IEEE International Conference on Consumer Technology-Europe (ICCT-Europe).

[34] Li, S. (2006). Trinocular spherical stereo. Proceedings of the IEEE/RSJ International Conference on Intelligent Robots and Systems (IROS), 4786–4791.

[35] Cheng, X., Zhong, Y., Harandi, M., Dai, Y., Chang, X., Li, H., Drummond, T., & Ge, Z. (2020). Hierarchical neural architecture search for deep stereo matching. Advances in Neural Information Processing Systems (NeurIPS).